\documentclass[lettersize,journal]{IEEEtran}
\usepackage{amsmath,amsfonts}
\usepackage{algorithmic}
\usepackage{algorithm}
\usepackage{array}
\usepackage[caption=false,font=normalsize,labelfont=sf,textfont=sf]{subfig}
\usepackage{textcomp}
\usepackage{stfloats}
\usepackage{url}
\usepackage{verbatim}
\usepackage{graphicx}
\usepackage{cite}
\usepackage[dvipsnames]{xcolor}
\usepackage[colorlinks=true,linkcolor=blue,citecolor=blue,urlcolor=blue]{hyperref}
\usepackage[nameinlink,noabbrev]{cleveref}
\usepackage{booktabs}
\usepackage{multirow}
\crefname{figure}{Fig.}{Figs.}
\Crefname{figure}{Figure}{Figures}
\crefname{equation}{Eq.}{Eqs.}
\Crefname{equation}{Equation}{Equations}
\crefname{table}{Table}{Tables}
\Crefname{table}{Table}{Tables}
\begin{document}

\title{Unified Unsupervised and Sparsely-Supervised 3D Object Detection by Semantic Pseudo-Labeling and Prototype Learning}

\author{Yushen He, Lei Zhao, and Weidong Chen, \IEEEmembership{Member, IEEE} % <-this % stops a space
\thanks{Weidong Chen is the corresponding author (e-mail: wdchen@sjtu.edu.cn).}% <-this % stops a space
\thanks{The authors are with School of Automation and Intelligent Sensing and Institute of Medical Robotics, Shanghai Jiao Tong University, and Key Laboratory of System Control and Information Processing, Ministry of Education of China, Shanghai 200240, China.}%
}

% The paper headers
% \markboth{Journal of \LaTeX\ Class Files,~Vol.~14, No.~8, August~2021}%
% {Shell \MakeLowercase{\textit{et al.}}: A Sample Article Using IEEEtran.cls for IEEE Journals}

% \IEEEpubid{0000--0000/00\$00.00~\copyright~2021 IEEE}
% Remember, if you use this you must call \IEEEpubidadjcol in the second
% column for its text to clear the IEEEpubid mark.

\maketitle

\begin{abstract}
3D object detection is essential for autonomous driving and robotic perception, yet its reliance on large-scale manually annotated data limits scalability and adaptability. To reduce annotation dependency, unsupervised and sparsely-supervised paradigms have emerged. However, they face intertwined challenges: low-quality pseudo-labels, unstable feature mining, and a lack of a unified training framework. This paper proposes SPL, a unified training framework for both unsupervised and sparsely-supervised 3D object detection via \underline{S}emantic \underline{P}seudo-labeling and prototype \underline{L}earning. SPL first generates high-quality pseudo-labels by integrating image semantics, point cloud geometry, and temporal cues, producing both 3D bounding boxes for dense objects and 3D point labels for sparse ones. These pseudo-labels are not used directly but as probabilistic priors within a novel, multi-stage prototype learning strategy. This strategy stabilizes feature representation learning through memory-based initialization and momentum-based prototype updating, effectively mining features from both labeled and unlabeled data. Extensive experiments on KITTI and nuScenes datasets demonstrate that SPL significantly outperforms state-of-the-art methods in both settings. Our work provides a robust and generalizable solution for learning 3D object detectors with minimal or no manual annotations. Our code is available at \url{https://github.com/TossherO/SPL}.
\end{abstract}


\section{Introduction}

\IEEEPARstart{3}{D} object detection is a critical perception task for applications like autonomous driving and robot navigation, aiming to locate objects within 3D space from sensor inputs such as LiDAR or cameras. While fully-supervised methods\cite{voxelrcnn,pvrcnn,bevfusion,cmt} have advanced significantly by leveraging large-scale annotated datasets (e.g., KITTI\cite{kitti}, nuScenes\cite{nuscenes}, Waymo\cite{waymo}), acquiring accurate 3D bounding box annotations remains costly and labor-intensive. This practical limitation has spurred research into two alternative paradigms: unsupervised and sparsely-supervised 3D object detection.

Unsupervised methods avoid human annotations entirely by generating 3D Bbox pseudo labels from data itself. Some methods\cite{oyster,cpd} exploit geometric features and commonsense priors in point clouds. Other methods utilize either temporal motion cues \cite{modest,liso,motal} or projected 2D image semantics \cite{fgr,annofreeod,dfu3d} to generate pseudo labels. A few methods\cite{lise,union} combine both motion and semantic information. Unsupervised methods primarily focus on improving the quality of generated pseudo labels to enhance detection accuracy.

Sparsely-supervised methods, in contrast, utilize a very limited set of human annotations — only a small subset of training samples are labeled, and each may contain only a single object annotation. These methods typically employ specialized training strategies, such as contrastive learning\cite{coin,cpdet3d}, to enable the model to learn from sparse supervision and generalize to unlabeled objects. An effective training strategy enables sparsely-supervised methods to maintain high performance despite using only sparse annotations.

\begin{figure}[t]
  \centering
  \includegraphics[width=8.5cm]{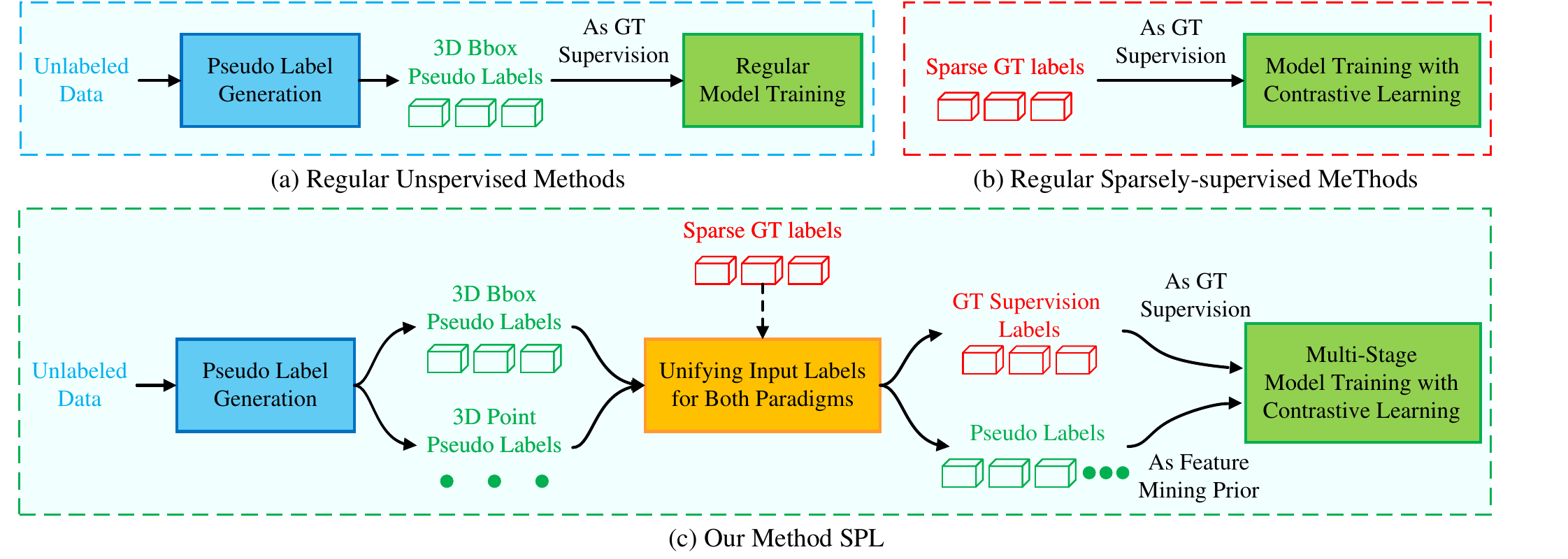}
  \caption{Comparison between representative unsupervised and sparsely-supervised 3D object detection pipelines and our unified SPL framework. Existing methods usually emphasize either pseudo-label generation or sparse-supervision feature learning, while SPL explicitly integrates semantic pseudo-labeling and prototype-based feature mining in a single training framework.}
  \label{fig:head}
\end{figure}

Despite their promise, both paradigms face fundamental and interrelated challenges:

(1) Low-quality pseudo labels in unsupervised learning. Motion-based methods\cite{modest,liso,motal} fail on stationary or slow-moving objects and struggle with class distinction. Methods\cite{fgr,annofreeod,dfu3d} relying on image semantics suffer from projection errors and background clutter, and generally do not utilize temporal information effectively. Hybrid approaches like LiSe\cite{lise} and UNION\cite{union} still inherit the respective limitations of each type of methods. Furthermore, for objects with sparse point clouds, these methods struggle to generate pseudo labels or produce severely mis-sized pseudo labels.

\begin{figure}[t]
  \centering
  \includegraphics[width=8.5cm]{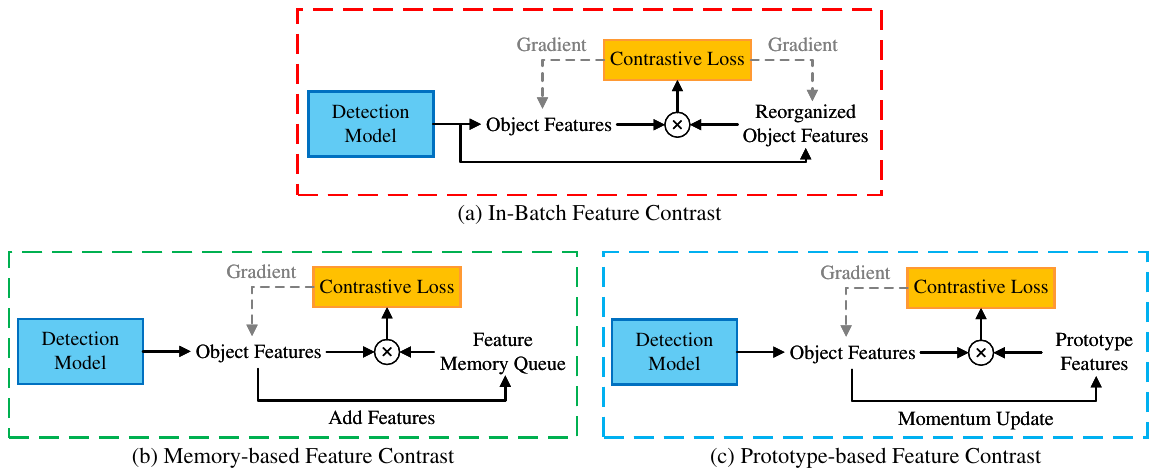}
  \caption{Comparison of contrastive learning strategies under sparse 3D supervision. The figure contrasts in-batch contrast, memory-queue contrast, and prototype-based contrast, and highlights their different trade-offs in sample diversity, feature consistency, and training stability.}
  \label{fig:constrasive_learning}
\end{figure}

(2) Unstable feature mining in sparsely-supervised learning. Current strategies based on contrastive learning exhibit significant limitations in sparse settings, as categorized in \cref{fig:constrasive_learning}: (a) In-batch feature contrast (e.g., CoIn \cite{coin}) uses only the limited object features available in a batch to construct pairs, leading to unstable training due to inadequate sample diversity. (b) Feature memory queue maintains a fixed-length queue for contrast. While mitigating sample scarcity, it introduces inconsistency as older stored features become outdated relative to the evolving model. (c) Prototype-based contrast (e.g., CPDet3D \cite{cpdet3d}) employs multiple prototypes each class to represent class features for stable comparison. Contrastive loss is computed between current object features and prototypes, and prototypes are updated via momentum using current features. However, its common random prototype initialization is detrimental to proper prototype learning. These inherent shortcomings in existing contrastive learning strategies result in sub-optimal feature discrimination and representation learning.

(3) Lack of a unified training framework adaptable to both paradigms. Existing approaches are narrowly designed for one setting and fail to leverage the synergy between pseudo label quality and feature mining. In unsupervised training, most methods \cite{fgr,modest,oyster,cpd,lise,liso,union} focus mainly on generating high-quality pseudo labels. A few others \cite{motal,annofreeod} incorporate auxiliary cues from pseudo labels to optimize loss computation but still do not engage in deep model representation learning. Conversely, in sparsely-supervised training, works like \cite{coin,cpdet3d} concentrate solely on designing feature mining strategies using the sparse ground truth. However, high-quality pseudo labels and robust feature mining are complementary and crucial for both settings.

To address these challenges, we propose SPL, a unified training framework centered on \textbf{S}emantic \textbf{P}seudo-labeling and prototype \textbf{L}earning, which is simultaneously adaptable to both unsupervised and sparsely-supervised 3D object detection. As shown in \cref{fig:head}, SPL first generates high-quality pseudo labels based on image semantics, then trains the model through a multi-stage strategy centered on prototype learning.

For Challenge 1, we introduce a high-quality pseudo label generation method. It begins with 3D instance segmentation derived from image semantics and point cloud projection. We then employ point cloud geometric features to resolve misassigned, missing, and overlapping points. For objects with low point density, we record them as 3D point pseudo labels. For others, we fit 3D Bboxes, refine them using temporal information. Thus it produces both high-quality 3D Bbox pseudo labels and point-level labels for sparse objects.

For Challenge 2, we design a multi-stage prototype learning strategy for stable feature mining. In Stage 1, we use a feature memory queue (\cref{fig:constrasive_learning}(b)) to gather diverse features and initialize prototypes via clustering. Stage 2 adopts a prototype-based strategy (\cref{fig:constrasive_learning}(c)), updating prototypes conservatively using only ground-truth features. Stage 3 further introduces pseudo heatmap priors for comprehensive feature mining, incorporating background contrast to enhance representation learning.

For Challenge 3, we unify the input supervision for both paradigms. We define two label types: ``GT Supervision Labels'' and ``Pseudo Labels''. For sparsely-supervised training, normal inputs are used. For unsupervised training, an evaluation score converts high-quality pseudo labels into ``GT Supervision Labels'', with the rest treated as ``Pseudo Labels''. Crucially, pseudo labels do not serve as direct supervision. Instead, they act as pseudo heatmap priors, together with prototypes, to guide the feature mining process alongside prototypes. This effectively couples pseudo-label information with representation learning.

Our contributions are summarized as follows:
\begin{itemize}
\item We propose SPL, a unified training framework based on Semantic Pseudo-Labeling and Prototype Learning, adaptable to both unsupervised and sparsely-supervised 3D object detection.
\item We introduce a high-quality pseudo label generation strategy that combines image semantics, point cloud geometry, and temporal information, producing not only high-quality 3D Bbox pseudo labels for dense objects but also 3D point pseudo labels for sparse objects.
\item We design a multi-stage training strategy centered on prototype learning, which unifies unsupervised and sparsely-supervised inputs via an evaluation score, couples pseudo labels with feature mining through pseudo heatmap priors, and stabilizes prototype initialization and updating while promoting deep representation learning.
\item Extensive experiments on KITTI and nuScenes datasets demonstrate that our method outperforms existing approaches in both unsupervised and sparsely-supervised 3D object detection tasks.
\end{itemize}

\section{Related Work}

\subsection{Fully-Supervised 3D Object Detection}

3D object detection primarily takes LiDAR point clouds as input. VoxelNet\cite{voxelnet} first uses neural network for this task by voxelizing point clouds, with subsequent works like SECOND\cite{second} and CenterPoint\cite{centerpoint} improving the model. Then some methods\cite{pointpillars} focus on achieving faster inference speeds, while others\cite{voxelrcnn,pvrcnn} enhance accuracy through two-stage frameworks. Additionally, some works\cite{smoke,caddn,petr} explore camera-only detection to reduce sensor costs, while multi-modal methods\cite{mvxnet,bevfusion,cmt,synet} fuse LiDAR and camera data for higher performance. Despite their strong performance, these approaches rely heavily on large-scale, accurately annotated 3D datasets, which limits their scalability and adaptability across different environments, sensor setups, and platforms.

\subsection{Unsupervised 3D Object Detection}

Early unsupervised methods\cite{nolearn1,nolearn2,nolearn3} produce detection results via non-learning strategies. More recent methods generally follow a two-stage pipeline: first generating pseudo 3D Bbox labels from unlabeled data, and then training a detector with these pseudo labels. Existing methods can be categorized based on their pseudo label generation mechanisms: (1) Geometry-based methods\cite{oyster,cpd} exploit geometric cues and common-sense priors from point clouds. (2) Motion-based methods\cite{modest,liso,motal} leverage motion cues across consecutive frames to identify moving objects. (3) Image-semantic-based methods\cite{fgr,annofreeod,dfu3d} project 3D point clouds onto 2D images to obtain semantic segmentation. (4) Hybrid methods\cite{lise,union} combine both motion and semantic information. However, the quality of generated pseudo labels remains a bottleneck. Moreover, these methods lack training strategies that encourage the learning of discriminative representations.

\subsection{Sparsely-Supervised 3D Object Detection}

SS3D\cite{ss3d} pioneers sparsely-supervised 3D detection by employing GT Sampling and self-training. CoIn\cite{coin} introduces contrastive learning using object features within mini-batches to mine features for unlabeled objects. HINTED\cite{hinted} extends CoIn with a mixed-feature augmentation strategy. CPDet3D\cite{cpdet3d} adopts a prototype-based contrastive learning for efficient feature representation. SP3D\cite{sp3d} attempts to enrich the sparse ground truth by generating additional pseudo labels. HASS\cite{hass} explores scene synthesis with a dynamic pseudo-database. However, the feature mining strategies of some methods — often based on contrastive learning — still face issues of training instability. Furthermore, they do not integrate pseudo labels into the feature mining process.

\subsection{Prototype-based Methods}

Prototype-based methods\cite{proto2d1,proto2d2,proto2d5} have been widely explored in 2D object detection, segmentation, and representation learning. In 3D detection, some works\cite{proto3d1,proto3d2,proto3d3} utilize geometric prototypes for tasks such as unsupervised and domain-adaptive detection. MoCo\cite{moco} introduces a momentum update mechanism for maintaining consistent feature queues in contrastive learning. CPDet3D\cite{cpdet3d} adapts ideas from MoCo and uses them for feature-level prototype learning in sparsely-supervised 3D detection. Our method draws inspiration from MoCo and CPDet3D.

\section{Method}

\begin{figure*}[t]
  \centering
  \includegraphics[width=15cm]{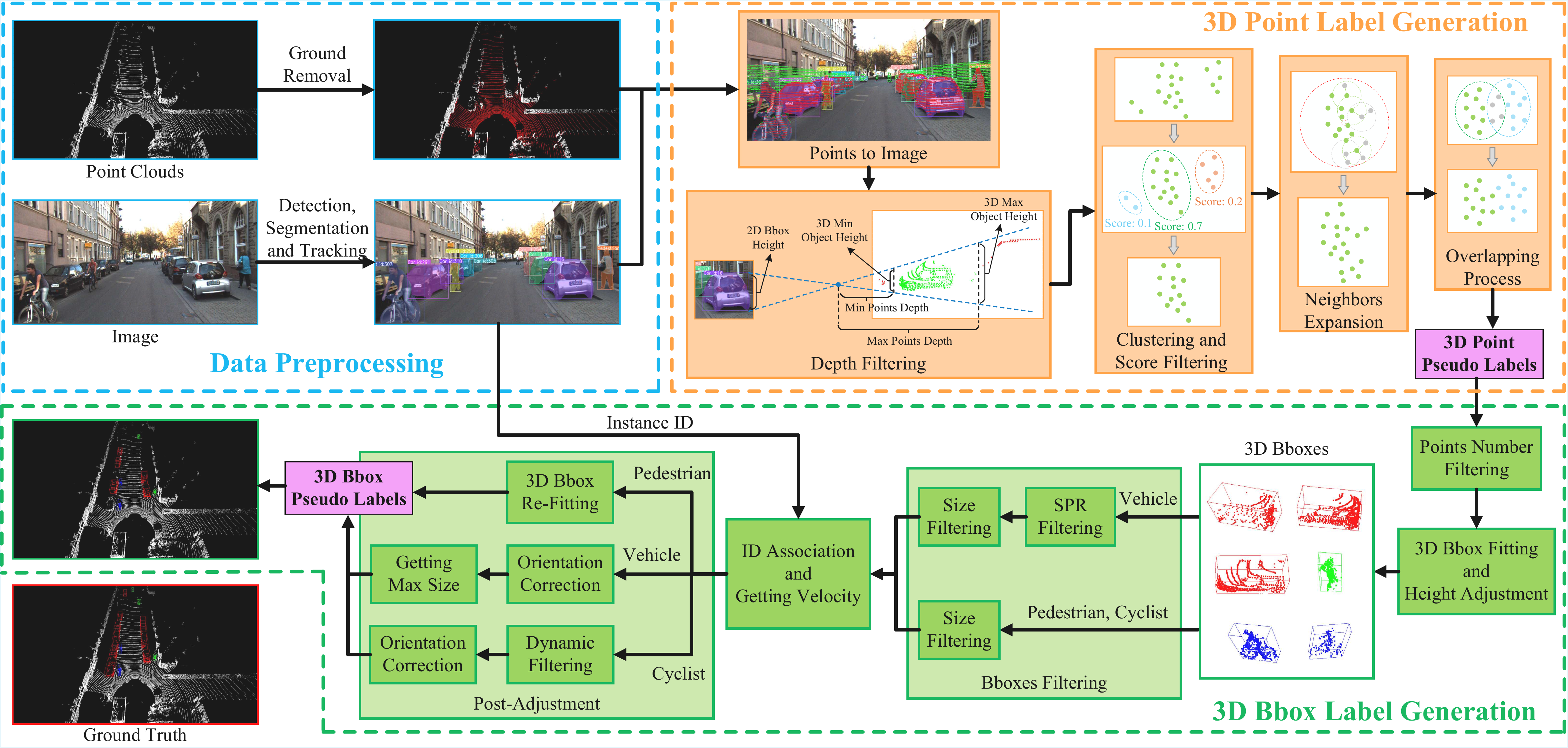}
  \caption{Overview of the proposed 3D pseudo-label generation pipeline. Starting from synchronized LiDAR frames and RGB images, the pipeline performs data preprocessing, 3D point label generation, and 3D Bbox label generation with temporal refinement, producing both high-quality 3D Bbox pseudo labels and 3D point pseudo labels for sparse objects.}
  \label{fig:pseudo_labels}
\end{figure*}

We propose SPL, a unified training framework that is simultaneously adaptable to both unsupervised and sparsely-supervised 3D object detection. SPL first generates high-quality 3D Bbox pseudo labels by integrating image semantics, point cloud geometry, and temporal cues, and additionally produces 3D point pseudo labels for objects with sparse point clouds. It then unifies the input supervision for both paradigms into two types: ``GT supervision labels'' and ``pseudo labels''. The training framework employs a prototype-based training strategy to train the 3D object detector, and adopts a multi-stage training scheme to achieve stable and effective representation learning.

We now present the details of our approach. In \cref{sec:pseudo_label_generation}, we describe our 3D pseudo label generation strategy. \cref{sec:prototype_based_training_strategy} introduces the prototype-based training strategy at the core of our framework. Finally, \cref{sec:multi_stage_training_pipeline} elaborates on the multi-stage training pipeline implemented in SPL.

\subsection{3D Pseudo Label Generation}\label{sec:pseudo_label_generation}

This section presents our method for generating high-quality 3D pseudo labels. As illustrated in \cref{fig:pseudo_labels}, the process takes point clouds and images of a continuous scene as input, and proceeds through three sequential stages: Data Preprocessing, 3D Point Label Generation, and 3D Bbox Label Generation, yielding 3D pseudo labels for each frame.

\subsubsection{Data Preprocessing}

We begin by preprocessing raw point clouds and images. For LiDAR point clouds, we aggregate multi-frame data using ego-vehicle poses and remove ground points via the Patchwork++ algorithm \cite{patchwork++}. For RGB images, to ensure efficiency, we employ the detector YOLOv12 \cite{yolov12} coupled with the tracker BoT-SORT \cite{botsort} to obtain per-frame object class labels $C^{t} = [c_{1}^{t}, c_{2}^{t}, \dots]$, 2D Bboxes $B^{t} = [b_{1}^{t}, b_{2}^{t}, \dots]$, instance segmentation masks $M^{t} = [m_{1}^{t}, m_{2}^{t}, \dots]$, and consistent cross-frame object IDs. This process is shown in \cref{eq:1}, where $I^{t}$ denotes the image at frame $t$. The object classes (Vehicle, Pedestrian, Cyclist) align with common autonomous driving benchmarks.

\begin{small}
\begin{equation}
  [C^{t},B^{t},M^{t},ID^{t}]=f_{det2d+track}(I^{t})
  \label{eq:1}
\end{equation}
\end{small}

\subsubsection{3D Point Label Generation}

This stage aims to generate reliable 3D point pseudo labels by fusing 2D image semantics with 3D point cloud geometry.

First, we project the ground-removed point cloud $Pc^{t}$ onto the image plane to obtain $Pc_{im}^{t}$. For each 2D object $i$ with associated $[c_{i}^{t}, b_{i}^{t}, m_{i}^{t}]$, we extract points in $Pc_{im}^{t}$ falling within its mask $m_{i}^{t}$ as the corresponding point cloud $pc_{i}^{t}$. The mask $m_{i}^{t}$ is dilated to include more points.

For each object category $c$, we define a real-world height range $[h_{c,\min}, h_{c,\max}]$. Given the pixel height $h_{i}^{t}$ of object $i$ in the image, the depth range $[d_{i,\min}^{t}, d_{i,\max}^{t}]$ of its point cloud is computed by \cref{eq:2}, where $f_{y}$ is the camera's vertical focal length. We filter out points in $pc_{i}^{t}$ that fall outside this depth range.

\begin{small}
\begin{equation}
  d_{i,min}^{t}=\frac{f_{y}\cdot h_{c,min}}{h_{i}^{t}}, \quad d_{i,max}^{t}=\frac{f_{y}\cdot h_{c,max}}{h_{i}^{t}}
  \label{eq:2}
\end{equation}
\end{small}

We then address misassigned, missing, and overlapping points in $pc_{i}^{t}$ through the following steps:

{\renewcommand{\labelenumi}{\alph{enumi}.}
\begin{enumerate}
  \item Removing misassigned points: We cluster $pc_{i}^{t}$ using DBSCAN \cite{dbscan} into subclusters $[pc_{i,1}^{t}, pc_{i,2}^{t}, \dots]$. A fitting score for each subcluster is computed by \cref{eq:3}, considering both the proportion of points within the image mask and the cluster size. The subcluster with the highest score is retained as the updated $pc_{i}^{t}$.
  \begin{small}
  \begin{equation}
    score(k)=\frac{|\{p \in pc_{i,k}^{t} | p \in m_{i}^{t}\}|}{|pc_{i,k}^{t}|} + \frac{|pc_{i,k}^{t}|}{\max_{j} |pc_{i,j}^{t}|}
    \label{eq:3}
  \end{equation}
  \end{small}
  \item Recovering missing points: Given a search radius $r_1$ and a maximum radius $r_2$, we iteratively add points within $r_1$ of the current $pc_{i}^{t}$ and within $r_2$ of $pc_{i}^{t}$'s initial centroid until no further points can be added.
  \item Resolving point ownership conflicts: For points claimed by multiple objects, a K-nearest neighbor majority voting algorithm \cite{knn} determines the final ownership based on the predominant object label in their neighborhood.
\end{enumerate}
}

After these steps, each object $i$ is associated with a refined point cloud $pc_{i}^{t}$. Its centroid $p_{i}^{t}$ is then recorded as the 3D point pseudo label.

\subsubsection{3D Bbox Label Generation}

Based on the 3D point pseudo labels, we generate 3D Bbox pseudo labels and refine them using temporal consistency.

First, for each object $i$, if the number of its corresponding points $|pc_{i}^{t}|$ exceeds a threshold, a 3D Bbox $b_{3d,i}^{t} = [x, y, z, l, w, h, \theta]$ is estimated via the L-shape fitting algorithm \cite{lshape}, with its height adjusted to ground contact. 3D Bboxes with unreasonable length, width, or height are filtered out. For Vehicle objects, we further compute the Surface Proximity Ratio (SPR), the proportion of points lying close to the box surface, and discard Bboxes with low SPR.

Next, we use the object IDs obtained during image preprocessing for cross-frame association. For object $i$ present in consecutive frames, its velocity $v_{i}^{t}$ is computed using the centroid displacement via \cref{eq:4}, where $\Delta p_{i}^{t-1} = p_{i}^{t} - p_{i}^{t-1}$, $\Delta p_{i}^{t} = p_{i}^{t+1} - p_{i}^{t}$, and $\Delta t_{t-1}$, $\Delta t_{t}$ are inter-frame time intervals.

\begin{small}
\begin{equation}
  v_{i}^{t} = \frac{\Delta p_{i}^{t-1} \cdot \Delta t_{t} + \Delta p_{i}^{t} \cdot \Delta t_{t-1}}{\Delta t_{t-1} + \Delta t_{t}}
  \label{eq:4}
\end{equation}
\end{small}

Finally, we integrate motion cues to refine the 3D Bboxes: 
(a) For Pedestrian objects, the 3D Bbox orientation is aligned with the velocity direction, and the Bbox is refitted accordingly. 
(b) For Vehicle and Cyclist objects, 3D Bboxes with an orientation deviating more than 90 degrees from the velocity direction are reversed. 
(c) To counteract undersized Bboxes due to partial scanning, vehicle dimensions are set to the maximum observed over consecutive frames. 
(d) Stationary Cyclist Bboxes are removed to avoid treating parked riders as valid detections.

In summary, high-quality 3D Bbox pseudo labels are produced for well-scanned objects, while 3D point pseudo labels are retained for sparse objects, ensuring their participation in subsequent training.

\subsection{Prototype-Based Training Strategy}\label{sec:prototype_based_training_strategy}

We build our prototype-based training strategy upon a conventional 3D detection architecture. We adopt CenterPoint\cite{centerpoint} as the baseline detector, modifying only its loss computation while leaving the inference network unchanged. Our strategy is also applicable to other detectors, such as the two-stage detector Voxel-RCNN\cite{voxelrcnn} and the multimodal detectors BEVFusion\cite{bevfusion}, provided they use heatmaps for classification loss calculation, as in CenterPoint.

\begin{figure*}[t]
  \centering
  \includegraphics[width=14cm]{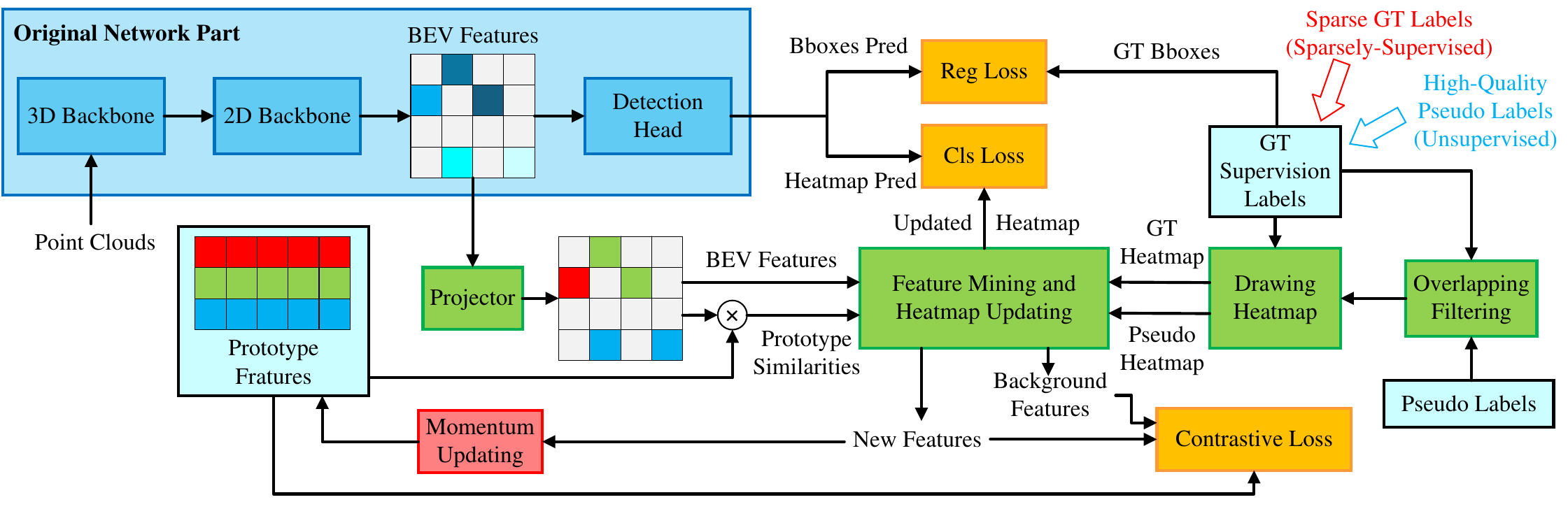}
  \caption{Overview of the prototype-based training strategy in SPL. The framework unifies supervision into GT supervision labels and pseudo labels, mines foreground/background features from BEV representations, and optimizes detector training with heatmap supervision and intra/inter-class contrastive objectives while updating prototypes with momentum.}
  \label{fig:train_framework}
\end{figure*}

As illustrated in \cref{fig:train_framework}, the core of our approach lies in maintaining and updating a set of prototypes $P \in \mathbb{R}^{C \times K \times D}$, where $C$, $K$, and $D$ denote the number of object classes, prototypes per class, and feature dimension respectively. During training, we encourage intermediate BEV features to align with corresponding prototypes while leveraging pseudo labels as additional guidance for feature mining. In the following, we detail our training strategy, covering Labels Processing, Feature Mining, Loss Function, and Prototype Update.

\subsubsection{Labels Processing}

We define two label types for both training paradigms: ``GT Supervision Labels'' and ``Pseudo Labels''. For sparsely-supervised training, sparse human annotations serve as ``GT Supervision Labels''. For unsupervised training, we convert high-quality 3D Bbox pseudo labels into ``GT Supervision Labels''. Quality is measured by high 2D-3D alignment (IoU between projected 3D Bbox and 2D mask) and evidence of dynamic motion. Both 3D Bbox and 3D point pseudo labels (after removing overlaps with GT Supervision Labels) are collected as ``Pseudo Labels''.

We generate a GT heatmap $H_{g}$ from GT Supervision Labels and a pseudo heatmap $H_{p}$ from Pseudo Labels using the standard heatmap generation procedure of CenterPoint. Since 3D point pseudo labels lack size, we assign them the minimum size of their respective class.

\subsubsection{Feature Mining}

\begin{figure*}[t]
  \centering
  \includegraphics[width=14cm]{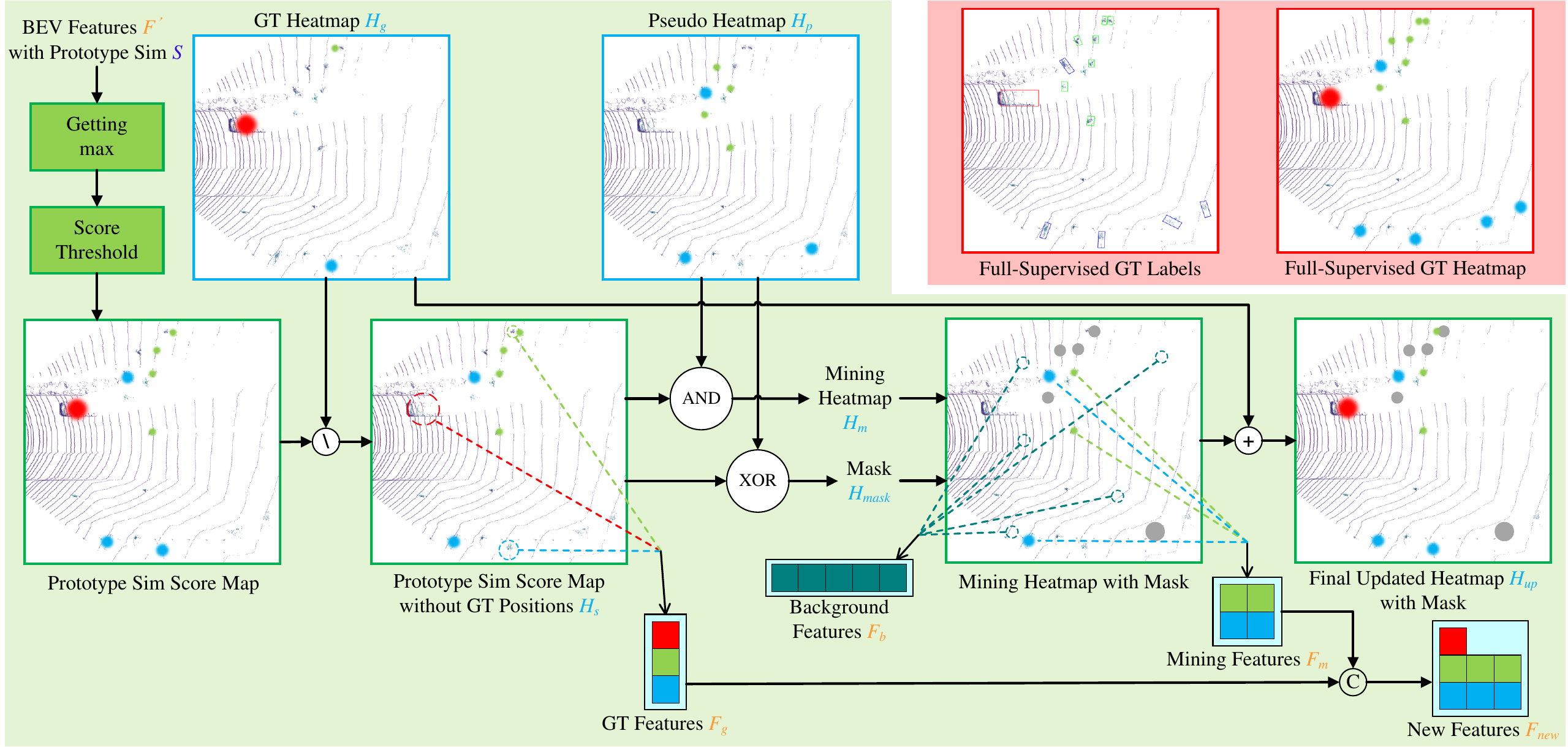}
  \caption{Feature mining process that fuses prototype similarity with pseudo-heatmap priors. Candidate regions are first selected by high prototype similarity and then cross-validated with pseudo labels to form reliable mined positives and masked ambiguous regions, which are subsequently used to construct contrastive foreground/background feature sets.}
  \label{fig:feature_mining}
\end{figure*}

The Feature Mining step identifies potential unlabeled objects by combining prototype similarity and pseudo heatmap priors. Given intermediate BEV features $F$, we project them through a small network $\text{Proj}(\cdot)$ consisting of a 2D convolution followed by L2 normalization, yielding $F^{'} = \text{Proj}(F) \in \mathbb{R}^{H \times W \times D}$, where $H$ and $W$ are the spatial dimensions of the feature map. Because both $F^{'}$ and the prototypes $P$ are normalized, we compute their cosine similarity $S \in \mathbb{R}^{H \times W \times C \times K}$ via dot product:

\begin{small}
\begin{equation}
  \begin{gathered}
    S(h,w,c,k) = F^{'}(h,w) \cdot P(c,k), \\
    h \in [1,H],~ w \in [1,W],~ c \in [1,C],~ k \in [1,K]
  \end{gathered}
  \label{eq:5}
\end{equation}
\end{small}

The following processing is shown in \cref{fig:feature_mining}. We reduce $S$ by taking the maximum similarity at each spatial location, producing $S^{'} \in \mathbb{R}^{H \times W}$ along with corresponding classes $C_{id}^{s} \in \mathbb{R}^{H \times W}$ and prototype indices $K_{id}^{s} \in \mathbb{R}^{H \times W}$:

\begin{small}
\begin{equation}
  \begin{gathered}
    S^{'}(h,w) = \max_{c,k} S(h,w,c,k) \\
    C_{id}^{s}(h,w), K_{id}^{s}(h,w) = \arg\max_{c,k} S(h,w,c,k)
  \end{gathered}
  \label{eq:6}
\end{equation}
\end{small}

We filter low-confidence locations by a fixed threshold $\tau_{s}$, and exclude areas already occupied by GT objects, yielding a similarity score map $H_{s}$:

\begin{small}
\begin{equation}
  H_{s}(h,w) =
  \begin{cases}
    S^{'}(h,w), & \text{if } S^{'}(h,w) > \tau_{s}
    \text{ and } H_{g}(h,w) = 0, \\
    0, & \text{otherwise}
  \end{cases}
  \label{eq:7}
\end{equation}
\end{small}

Then we fuse $H_{s}$ with the pseudo heatmap $H_{p}$ using two rules: (1) positions where both $H_{p}$ and $H_{s}$ are positive and class predictions agree receive positive supervision; (2) positions where either $H_{p}$ or $H_{s}$ is positive (but not both) are masked out from negative supervision to avoid suppressing potentially correct predictions. Formally, the mining heatmap $H_{m}$ (positive signals) and the mask $H_{mask}$ (ambiguous regions) are obtained by \cref{eq:8} and \cref{eq:9}. The final heatmap used for classification supervision is then $H_{up} = H_{g} + H_{m}$.

\begin{small}
\begin{equation}
  H_{m}(h,w) =
  \begin{cases}
    H_{s}(h,w), & \text{if } H_{p}(h,w) > 0 \text{ and } H_{s}(h,w) > 0 \\
    &\text{and } C_{id}^{s}(h,w) = C_{id}^{p}(h,w), \\
    0, & \text{otherwise}
  \end{cases}
  \label{eq:8}
\end{equation}
\end{small}

\begin{small}
\begin{equation}
  H_{mask}(h,w) =
  \begin{cases}
    0, & \text{if } H_{m}(h,w) = 0 \text{ and } \\
    &H_{p}(h,w) + H_{s}(h,w) > 0, \\
    1, & \text{otherwise}
  \end{cases}
  \label{eq:9}
\end{equation}
\end{small}

Furthermore, we extract feature vectors at the locations marked by $H_{g}$ and $H_{m}$ to form the set of foreground features $F_{new} = [F_{g}; F_{m}]$, where

\begin{small}
\begin{equation}
  \begin{gathered}
      F_{g} = \{F^{'}(h,w) \mid H_{g}(h,w) > 0\}, \\
      F_{m} = \{F^{'}(h,w) \mid H_{m}(h,w) > 0\}
  \end{gathered}
  \label{eq:10}
\end{equation}
\end{small}

We also collect the class indices $C_{id}^{new}$ and prototype indices $K_{id}^{new}$ of $F_{new}$. For $F_{g}$ part, the class indices are derived from the GT labels, and the prototype indices are the ones with the highest similarity. For $F_{m}$ part, both indices are obtained from \cref{eq:6}.

To strengthen the contrastive learning, we randomly sample a set of background features $F_{bg}$ by \cref{eq:11}, where $|F_{bg}| = C \times K$. 

\begin{small}
\begin{equation}
  \begin{gathered}
    F_{bg} = Sample(\{F^{'}(h,w) \mid (h,w) \in \Omega\}) \\
    \Omega = \{(h,w) \mid H_{g}(h,w)=0, H_{p}(h,w)=0, H_{s}(h,w)=0\}
  \end{gathered}
  \label{eq:11}
\end{equation}
\end{small}

\subsubsection{Loss Function}

The overall training loss comprises four components:

\begin{small}
\begin{equation}
  L = L_{reg} + L_{cls} + \lambda_{1} L_{con, intra} + \lambda_{2} L_{con, inter}
  \label{eq:12}
\end{equation}
\end{small}

The regression loss $L_{reg}$ follows CenterPoint's original formulation, using only the sparse ``GT Supervision Labels''. We find that even with few GT Bboxes, the model learns to predict reasonable 3D Bboxes.

The classification loss $L_{cls}$ is adapted from CenterPoint: the target heatmap is replaced by $H_{up}$, and the loss is computed only where  $H_{mask} = 1$ (ambiguous regions excluded).

The intra-class contrastive loss $L_{con, intra}$ is defined in \cref{eq:13}, where $\tau_{t}$ is the temperature parameter, which pulls each foreground feature toward its assigned prototype while pushing it away from other prototypes of the same class. 

\begin{footnotesize}
\begin{equation}
  \begin{gathered}
    L_{con, intra} = -\frac{1}{|F_{new}|} \sum_{i=1}^{|F_{new}|}
    \log \frac{\exp(F_{new, i} \cdot P(C_{id, i}^{new}, K_{id, i}^{new}) / \tau_t)}{\sum_{k=1}^{K} \exp(F_{new, i} \cdot P(C_{id, i}^{new}, k) / \tau_t)} \\
  \end{gathered}
  \label{eq:13}
\end{equation}
\end{footnotesize}

The inter-class contrastive loss $L_{con, inter}$ is given by \cref{eq:14}, which enhances discrimination across different categories and between foreground and background.

\begin{small}
\begin{equation}
  \begin{gathered}
    L_{con, inter} = -\frac{1}{|F_{new}|} \sum_{i=1}^{|F_{new}|}
    \log \frac{S_{self}}{S_{proto} + S_{bg}}, \text{where } \\
    \left\{
    \begin{array}{l}
      S_{self} = \exp(F_{new, i} \cdot P(C_{id, i}^{new}, K_{id, i}^{new}) / \tau_t), \\
      S_{proto} = S_{self} + \sum_{c \neq C_{id, i}^{new}} \sum_{k=1}^{K} \exp(F_{new, i} \cdot P(c,k) / \tau_t), \\
      S_{bg} = \sum_{j=1}^{|F_{bg}|} \exp(F_{new, i} \cdot F_{bg,j} / \tau_t)
    \end{array}
    \right.
  \end{gathered}
  \label{eq:14}
\end{equation}
\end{small}

\subsubsection{Prototype Updating}

We update the prototypes using a momentum-based approach inspired by MoCo \cite{moco}. For each prototype $P(c,k)$, we collect all features in $F_{new}$ assigned to it, denoted as $F_{c,k} = \{F_{new, i} | C_{id, i}^{new} = c, K_{id, i}^{new} = k\}$. If $F_{c,k}$ is non-empty, we compute its mean feature vector $\bar{F}_{c,k}$ and update the prototype via \cref{eq:15}, where $\alpha$ is the momentum coefficient.

\begin{small}
\begin{equation}
  \begin{gathered}
    P(c,k) \leftarrow Norm_{L2}(\alpha P(c,k) + (1 - \alpha) \bar{F}_{c,k}) \\
    \bar{F}_{c,k} = \frac{1}{|F_{c,k}|} \sum_{f \in F_{c,k}} f
  \end{gathered}
  \label{eq:15}
\end{equation}
\end{small}

In summary, our prototype-based training strategy effectively integrates both GT supervision and pseudo labels to guide feature mining and representation learning. The loss functions encourage discriminative feature learning, while the momentum-based prototype updating ensures stable and consistent prototype representations throughout training.

\subsection{Multi-Stage Training Pipeline}\label{sec:multi_stage_training_pipeline}

\begin{figure}[t]
  \centering
  \includegraphics[width=8cm]{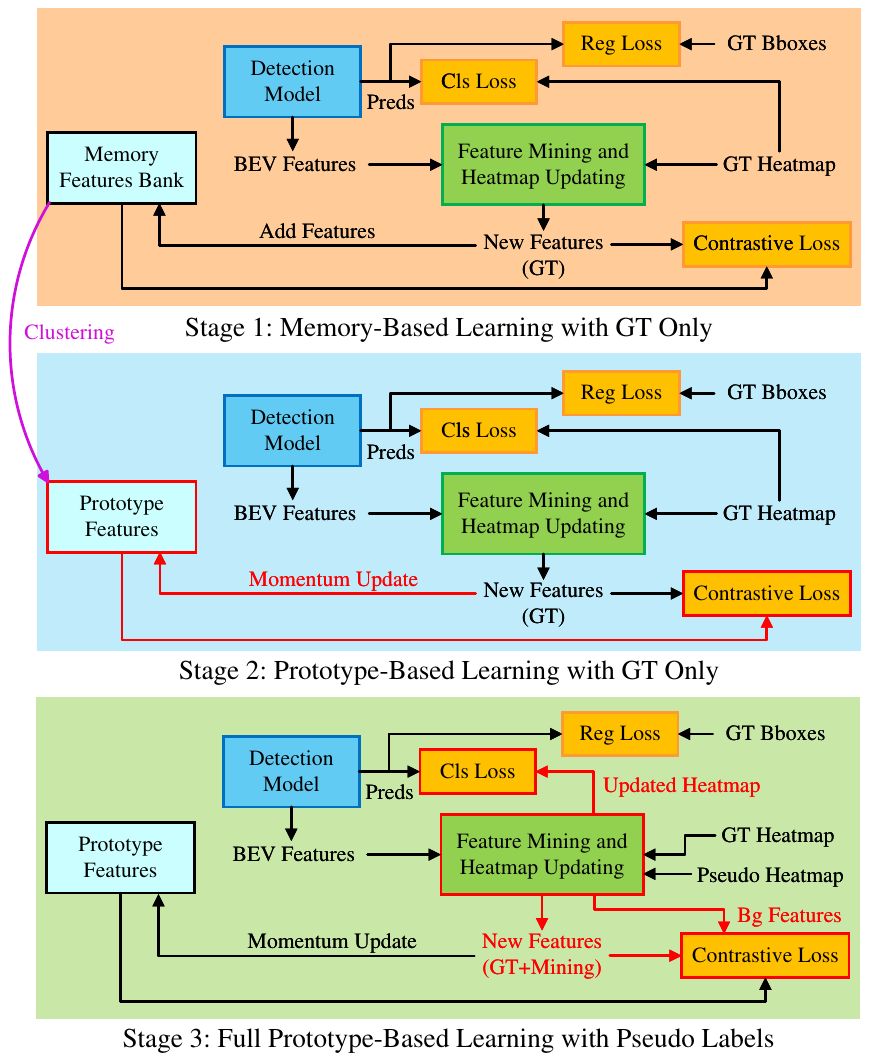}
  \caption{Overview of the three-stage training pipeline used to stabilize prototype learning. Stage 1 performs memory-based contrastive pretraining for robust prototype initialization, Stage 2 refines prototypes with GT-only supervision, and Stage 3 enables the full SPL strategy with pseudo-label-guided feature mining.}
  \label{fig:multi_stage}
\end{figure}

Direct application of the prototype-based training strategy often leads to unstable prototype learning due to poor initialization and the noisy nature of pseudo labels. To address this, we adopt a three-stage training pipeline that progressively transitions from memory-based feature collection to full prototype-based learning. As shown in \cref{fig:multi_stage}, each stage introduces increasingly complex components while maintaining training stability.

\subsubsection{Stage 1, Memory-Based Learning with GT Only}

This stage uses only ``GT Supervision Labels''. We adopt a memory-based contrastive learning strategy, maintaining a memory feature bank $M_{c}$ for each class $c$. During training, we extract features $F_{g}$ from GT-labeled objects and compute contrastive loss against stored features in $M_{c}$. For each class $c$, the memory bank is updated by enqueuing new features and dequeuing the oldest ones. The contrastive loss in this stage is computed as shown in \cref{eq:16}, where for each feature $F_{g, i}$, $m_{i}^{+}$ is the most similar positive feature from the same class in the memory bank, and $M_{i}^{-}$ contains negative features from other classes. This pulls features of the same class closer while pushing apart features from different classes.

\begin{scriptsize}
\begin{equation}
    L_{con} = -\frac{1}{|F_{g}|} \sum_{i=1}^{|F_{g}|}
    \log \frac{\exp(F_{g,i} \cdot m_i^+ / \tau_t)}
    {\exp(F_{g,i} \cdot m_i^+ / \tau_t) + \sum_{m_i^- \in M_i^-} \exp(F_{g,i} \cdot m_i^- / \tau_t)}
  \label{eq:16}
\end{equation}
\end{scriptsize}

At the end of Stage 1, we perform K-means clustering on the accumulated memory features of each class to initialize the prototypes, providing a robust starting point for subsequent stages.

\subsubsection{Stage 2, Prototype-Based Learning with GT Only}

With prototypes initialized, we now switch to the prototype-based learning strategy while still using only ``GT Supervision Labels''. The training process in this stage differs from the full strategy described in \cref{sec:prototype_based_training_strategy} in several aspects: 
(1) The classification loss $L_{cls}$ is computed in the conventional manner, using only the GT heatmap $H_{g}$ as the target.
(2) For contrastive learning, only foreground features from GT objects $F_{new} = F_{g}$ are used. Neither mined features $F_{m}$ nor background features $F_{bg}$ participate in the contrastive loss computation.
(3) Prototype updating uses only the GT foreground features $F_{g}$, without influence from pseudo-labeled or mined regions.
This stage focuses on stabilizing the initialized prototypes and learning robust feature representations before introducing pseudo labels in the next stage.

\subsubsection{Stage 3, Full Prototype-Based Learning with Pseudo Labels}

In this final stage, we activate the complete prototype-based training strategy from \cref{sec:prototype_based_training_strategy}, incorporating both ``GT Supervision Labels'' and ``Pseudo Labels'' for comprehensive feature mining and prototype learning.

\section{Experiments}

\begin{table*}[t]
  \caption{Comparison on Sparsely-Supervised 3D Object Detection on KITTI val set.}
  \label{tab:kitti_sparse}
  \centering
  \resizebox{0.7\textwidth}{!}{
  \begin{tabular}{@{}ccccccccccc@{}}
  \toprule
  \multirow{2}{*}{Annotation Rate} & \multirow{2}{*}{Method} & \multicolumn{3}{c}{\begin{tabular}[c]{@{}c@{}}Car\\ AP @ 3D-IoU 0.7\end{tabular}} & \multicolumn{3}{c}{\begin{tabular}[c]{@{}c@{}}Pedestrian\\ AP @ 3D-IoU 0.5 \end{tabular}} & \multicolumn{3}{c}{\begin{tabular}[c]{@{}c@{}}Cyclist\\ AP @ 3D-IoU 0.5\end{tabular}} \\ \cmidrule(l){3-11}
                                   &                         & ~Easy~        & ~Mod.~        & ~Hard~        & ~Easy~        & ~Mod.~        & ~Hard~        & ~Easy~        & ~Mod.~        & ~Hard~        \\ \midrule
  100\%                            & Voxel-RCNN              & 92.3          & 84.9          & 82.6          & 69.6          & 63.0          & 58.6          & 88.7          & 72.5          & 68.2          \\ \midrule
  \multirow{5}{*}{2\%}             & Voxel-RCNN              & 70.5          & 54.9          & 44.8          & 42.6          & 38.5          & 32.1          & 73.3          & 47.8          & 43.2          \\
                                   & CoIn                    & 89.1          & 70.2          & 55.6          & 50.8          & 45.2          & 39.6          & 80.2          & 52.3          & 48.6          \\
                                   & CoIn++                  & \textbf{92.0} & 79.5          & 71.5          & 46.7          & 36.1          & 31.2          & 82.0          & 58.4          & 54.6          \\
                                   & SP3D                    & 91.3          & 80.5          & 74.0          & 67.4          & 58.7          & 50.9          & \textbf{92.5} & \textbf{73.1} & 68.3          \\
                                   & SPL (Ours)              & 91.8          & \textbf{82.2} & \textbf{79.4} & \textbf{69.5} & \textbf{63.2} & \textbf{56.7} & 91.8          & 73.0          & \textbf{68.5} \\ \bottomrule
  \end{tabular}
  }
\end{table*}

\begin{table*}[t]
  \caption{Comparison on Sparsely-Supervised 3D Object Detection on nuScenes val set.}
  \label{tab:nuscenes_sparse}
  \centering
  \resizebox{0.8\textwidth}{!}{
  \begin{tabular}{@{}cccccccccccccc@{}}
  \toprule
  Annotation Rate       & Method      & mAP            & NDS            & Car            & Truck          & C.V.           & Bus            & Trailer        & Barrier        & Motor.        & Bike          & Ped.           & T.C.           \\ \midrule
  100\%                 & CenterPoint & ~56.11~        & ~64.61~        & ~84.63~        & ~52.87~        & ~16.40~        & ~66.85~        & ~36.85~        & ~65.79~        & ~53.97~       & ~35.39~       & ~83.46~        & ~64.91~        \\ \midrule
  \multirow{3}{*}{10\%} & CenterPoint & 8.09           & 25.77          & 24.62          & 2.84           & 0.00           & 15.66          & 0.00           & 4.07           & 3.33          & 0.29          & 25.11          & 4.96           \\
                        & CoIn        & 12.47          & 33.79          & 38.70          & 6.85           & 0.00           & 20.67          & 7.81           & 11.51          & 2.85          & \textbf{3.36} & 34.85          & 8.50           \\
                        & SPL (Ours)  & \textbf{38.70} & \textbf{48.96} & \textbf{72.41} & \textbf{42.03} & \textbf{21.02} & \textbf{59.11} & \textbf{23.29} & \textbf{47.45} & \textbf{3.94} & 0.00          & \textbf{70.51} & \textbf{47.22} \\ \bottomrule
  \end{tabular}
  }
\end{table*}

We evaluate the effectiveness of our SPL framework on KITTI\cite{kitti} and nuScenes\cite{nuscenes} datasets under both unsupervised and sparsely-supervised 3D object detection settings. We compare SPL with recent state-of-the-art methods and conduct ablation studies to validate each component.

\subsection{Datasets and Evaluation Metrics}

\subsubsection{KITTI Dataset}

The KITTI 3D object detection dataset\cite{kitti} contains 7,481 training samples. Following common practice, we split them into 3,712 for training and 3,769 for validation. For sparsely-supervised experiments, we adopt the same setting as CoIn\cite{coin}: only one object is annotated in 10\% of the scenes, resulting in approximately 2\% of the total annotations. For unsupervised experiments, we use all training samples without any 3D Bbox annotations. Evaluation uses the official Average Precision (AP) at 40 recall positions (R40) for the Car, Pedestrian, and Cyclist classes.

\subsubsection{nuScenes Dataset}

The nuScenes dataset\cite{nuscenes} is a large-scale multimodal dataset containing 1,000 driving scenes, split into 700 for training, 150 for validation, and 150 for testing. We use the training set (about 28,000 annotated keyframes) for training and the validation set (about 6,000 keyframes) for evaluation. For sparsely-supervised experiments, we follow CoIn\cite{coin} and annotate one object per keyframe, resulting in approximately 10\% of total annotations. For unsupervised experiments, we consider three major object classes: Vehicle (car, truck, bus, trailer, construction vehicle), Pedestrian, and Cyclist (motorcycle, bicycle). Evaluation uses the official mean Average Precision (mAP) and nuScenes Detection Score (NDS).

\subsection{Implementation Details}

\subsubsection{Pseudo-Label Generation}

We use YOLOv12\cite{yolov12} and BoT-SORT\cite{botsort} for 2D detection and tracking. For DBSCAN\cite{dbscan}, the parameter min-samples is 4, and the $\epsilon$ is set to 0.5m, 0.3m, and 0.3m for Vehicle, Pedestrian, and Cyclist. The search radius $r_1$ is set to 0.25m, 0.15m, and 0.15m for Vehicle, Pedestrian, and Cyclist, while the maximum radius $r_2 = 4 r_1$. For Surface Proximity Ratio (SPR), the minimum distance to the box surface is 0.2m, and the SPR threshold is set to 0.8. When converting 3D Bbox pseudo labels to ``GT Supervision Labels'' for unsupervised training, the IoU threshold for 2D-3D alignment is 0.4.

\subsubsection{Network and Training}

We implement our framework on two base detectors: CenterPoint\cite{centerpoint} and Voxel-RCNN\cite{voxelrcnn}. The Voxel-RCNN uses the same one-stage detection head as CenterPoint to apply our prototype-based training strategy, while retaining its original two-stage refinement module and loss functions unchanged. For KITTI experiments, we use Voxel-RCNN as the detector; for nuScenes, we use CenterPoint.

We set the number of prototypes per class $K=5$, feature dimension $D=64$, similarity threshold $\tau_{s}=0.9$, temperature $\tau_{t}=1.0$, and momentum coefficient $\alpha=0.9$. The memory bank size in Stage 1 is set to 1000. The loss weights $\lambda_{1}$ and $\lambda_{2}$ for the intra-class and inter-class contrastive losses are set to 0.5 and 1.0. The model is trained on 4 NVIDIA RTX 3090 GPUs. We use the Adam optimizer and the cosine annealing schedule with an initial learning rate of $3\times10^{-3}$ for CenterPoint and $1\times10^{-2}$ for Voxel-RCNN. We apply GT Sampling augmentation during training. For KITTI, the total training epochs are 80 (Stage 1: 10, Stage 2: 10, Stage 3: 60). For nuScenes, the total training epochs are 30 (Stage 1: 5, Stage 2: 5, Stage 3: 20). To further boost performance, we employ a self-training strategy: after training the model with the above pipeline, we use it to generate pseudo labels on the entire training set, and then retrain the detector using these pseudo labels as additional supervision.

For more implementation details on pseudo-label generation and training, please refer to the supplementary material.

\subsection{Comparison with State-of-the-Art Methods}

\subsubsection{Comparison on Sparsely-Supervised 3D Object Detection}

On KITTI (\cref{tab:kitti_sparse}), we compare SPL with CoIn\cite{coin} and SP3D\cite{sp3d} under the same 2\% annotation setting. All methods use Voxel-RCNN as the base detector. SPL achieves the best performance on most metrics, surpassing SP3D by 2.1\% on mean AP across all classes and difficulty levels. On nuScenes (\cref{tab:nuscenes_sparse}), we compare SPL with CoIn\cite{coin} under the same 10\% annotation setting, using CenterPoint as the base detector. SPL significantly outperforms CoIn by 26.23\% in mAP and 15.17\% in NDS. These results validate the superiority of our SPL framework in sparsely-supervised 3D object detection across different datasets and detectors.

\subsubsection{Comparison on Unsupervised 3D Object Detection}

\begin{table*}[t]
  \caption{Comparison on Unsupervised 3D Object Detection on KITTI val set.}
  \label{tab:kitti_unsup}
  \centering
  \resizebox{0.7\textwidth}{!}{
  \begin{tabular}{@{}ccccccccccc@{}}
  \toprule
  \multirow{2}{*}{~Train Set~} & \multirow{2}{*}{~Method~} & \multicolumn{3}{c}{\begin{tabular}[c]{@{}c@{}}Car\\ AP @ 3D-IoU 0.5 \end{tabular}} & \multicolumn{3}{c}{\begin{tabular}[c]{@{}c@{}}Pedestrian\\ AP @ 3D-IoU 0.5 \end{tabular}} & \multicolumn{3}{c}{\begin{tabular}[c]{@{}c@{}}Cyclist\\ AP @ 3D-IoU 0.5\end{tabular}} \\ \cmidrule(l){3-11} 
                               &                           & ~Easy~        & ~Mod.~        & ~Hard~        & ~Easy~        & ~Mod.~        & ~Hard~        & ~Easy~        & ~Mod.~        & ~Hard~        \\ \midrule
  \multirow{4}{*}{Waymo}       & MODEST                    & 47.6          & 33.4          & 30.6          & 1.3           & 2.2           & 2.3           & 0.1           & 0.0           & 0.0           \\
                               & OYSTER                    & 65.3          & 54.8          & 43.6          & 3.0           & 3.0           & 3.0           & 1.7           & 1.8           & 1.9           \\
                               & CPD                       & 90.9          & 81.0          & 79.8          & 17.1          & 15.2          & 14.2          & 11.1          & 7.3           & 6.5           \\
                               & Motal                     & \textbf{96.2} & \textbf{87.6} & \textbf{85.8} & 37.9          & 33.4          & 31.1          & 56.3          & 37.8          & 35.5          \\ \midrule
  \multirow{3}{*}{KITTI}       & OYSTER                    & 43.7          & 34.5          & 31.2          & 0.0           & 0.0           & 0.0           & 0.0           & 0.0           & 0.0           \\
                               & LISO                      & 62.4          & 53.7          & 45.6          & 13.4          & 10.8          & 8.2           & 20.4          & 13.7          & 10.3          \\
                               & SPL (Ours)                & 93.3          & 83.1          & 75.9          & \textbf{46.1} & \textbf{40.5} & \textbf{34.0} & \textbf{67.1} & \textbf{41.9} & \textbf{36.7} \\ \bottomrule
  \end{tabular}
  }
\end{table*}

\begin{table}[t]
  \caption{Comparison on Unsupervised 3D Object Detection on nuScenes val set.}
  \label{tab:nuscenes_unsup}
  \centering
  \resizebox{0.48\textwidth}{!}{
  \begin{tabular}{@{}ccccccc@{}}
  \toprule
  Method       & Detector       & mAP           & NDS           & Vehicle       & Pedestrian    & Cyclist      \\ \midrule
  UNION        & CenterPoint    & 25.1          & 24.4          & 31.0          & 44.2          & 0.0          \\
  AnnofreeOD   & Voxel-NeXt     & 34.4          & 36.6          & 44.1          & 51.1          & \textbf{7.9} \\
  SPL (Ours)   & CenterPoint    & \textbf{38.3} & \textbf{40.6} & \textbf{48.1} & \textbf{59.3} & 7.2          \\ \bottomrule
  \end{tabular}
  }
\end{table}

On KITTI (\cref{tab:kitti_unsup}), we compare SPL with several recent unsupervised 3D detection methods, including MODEST\cite{modest}, OYSTER\cite{oyster}, CPD\cite{cpd}, Motal\cite{motal}, and LISO\cite{liso}. All methods use Voxel-RCNN as the base detector. Since KITTI has limited data, many prior works train their models on the larger Waymo dataset\cite{waymo} and evaluate on KITTI. In contrast, we train SPL directly on the KITTI training set without external data. Despite this, SPL outperforms all compared methods on most metrics, achieving significant improvements in AP for Pedestrian and Cyclist classes. On nuScenes (\cref{tab:nuscenes_unsup}), we compare SPL with UNION\cite{union} and AnnofreeOD\cite{annofreeod}, using CenterPoint and Voxel-NeXt\cite{voxelnext} as base detectors respectively. Due to the limitations of semantic-based pseudo-label generation, which cannot cover classes like Barrier and Traffic Cone, we only evaluate Vehicle, Pedestrian, and Cyclist classes. SPL achieves the best performance in both mAP and NDS. These results demonstrate the effectiveness of our SPL framework in unsupervised 3D object detection across different datasets and detectors.

For visualizations of detection results and the pseudo labels generated by our method, please refer to the supplementary material.

\subsection{Ablation Studies}

\subsubsection{Ablation on Pseudo Label Generation Strategy}

\begin{table*}[t]
  \caption{Ablation studies on Pseudo Label Generation Strategy on KITTI.}
  \label{tab:ablation_pseudo_label}
  \centering
  \resizebox{0.7\textwidth}{!}{
  \begin{tabular}{cccccccc}
  \toprule
  \multicolumn{2}{c}{\multirow{2}{*}{Pseudo Label Generation Method}} & \multicolumn{3}{c}{Precision}                 & \multicolumn{3}{c}{Recall}                    \\ \cmidrule(l){3-8}
  \multicolumn{2}{c}{}                                                 & Car           & Pedestrian    & Cyclist       & Car           & Pedestrian    & Cyclist       \\ \midrule
  \multicolumn{2}{c}{LISO}                                             & ~~~61.4~~~    & ~~~40.3~~~    & ~~~54.7~~~    & ~~~24.7~~~    & ~~~8.1~~~~    & ~~~17.4~~~    \\
  \multicolumn{2}{c}{UNION}                                            & 84.5          & 67.1          & 64.5          & 54.1          & 43.7          & 41.8          \\ \midrule
  \multirow{4}{*}{SPL~~~}   & (a) without 3D point pseudo labels       & \textbf{93.4} & \textbf{84.5} & \textbf{90.7} & 34.7          & 21.5          & 28.6          \\
                            & (b) with 3D point pseudo labels          & 82.6          & 64.3          & 76.1          & \textbf{67.4} & \textbf{54.9} & \textbf{61.7} \\
                            & (c) without object points refinement     & 70.6          & 57.6          & 67.8          & 59.3          & 47.1          & 53.7          \\
                            & (d) without 3D Bboxes refinement         & 73.4          & 59.0          & 70.9          & 62.4          & 51.7          & 57.4          \\ \bottomrule
  \end{tabular}
  }
\end{table*}

We first conduct ablation studies on the pseudo label generation strategy using the KITTI dataset. Since pseudo labels are not directly used for supervision in our SPL framework, we evaluate their quality using precision and recall metrics instead of detection performance. As shown in \cref{tab:ablation_pseudo_label}, we compare our full pseudo label generation method with LISO\cite{liso} and UNION\cite{union}. Our method achieves significantly higher recall while maintaining competitive precision. We further ablate key components of our strategy:
(a) using only 3D Bbox pseudo labels without 3D point pseudo labels results in high precision but low recall; 
(b) incorporating 3D point pseudo labels (set to the average size of the respective class) greatly improves recall at the cost of some precision;
(c) removing the refinement of object points (addressing misassigned, missing, and overlapping points) degrades both precision and recall;
(d) removing the refinement of 3D Bboxes with temporal information also leads to performance drops. These results validate the effectiveness of each component in our pseudo label generation strategy. 
The 3D Bbox pseudo labels can provide accurate supervision, while the 3D point pseudo labels enhance coverage and diversity for feature learning.

\subsubsection{Ablation on Training Strategy}

\begin{table}[t]
  \caption{Ablation studies on Prototype-Based Training Strategy on KITTI val set.}
  \label{tab:ablation_prototype_training}
  \centering
  \resizebox{0.48\textwidth}{!}{
  \begin{tabular}{@{}ccc|ccc@{}}
  \toprule
  \begin{tabular}[c]{@{}c@{}}Pseudo\\ ~~heatmap~~\end{tabular} & \begin{tabular}[c]{@{}c@{}}~~Prototype sim~~\\ score map\end{tabular} & \begin{tabular}[c]{@{}c@{}}Contrastive\\ loss\end{tabular} & \begin{tabular}[c]{@{}c@{}}Car\\ (AP @ 3D-IoU 0.7)\end{tabular} & \begin{tabular}[c]{@{}c@{}}Pedestrian\\ (AP @ 3D-IoU 0.5)\end{tabular} & \begin{tabular}[c]{@{}c@{}}Cyclist\\ (AP @ 3D-IoU 0.5)\end{tabular} \\ \midrule
                                                               &                                                                       &                                                            & 60.2                                                            & 38.4                                                                   & 52.1                                                                \\
  \checkmark                                                   &                                                                       &                                                            & 74.1                                                            & 47.6                                                                   & 61.3                                                                \\
                                                               & \checkmark                                                            &                                                            & 71.7                                                            & 45.7                                                                   & 57.0                                                                \\
                                                               & \checkmark                                                            & \checkmark                                                 & 77.9                                                            & 52.7                                                                   & 69.2                                                                \\
  \checkmark                                                   & \checkmark                                                            & \checkmark                                                 & \textbf{85.6}                                                   & \textbf{64.1}                                                          & \textbf{81.2}                                                       \\ \bottomrule
  \end{tabular}
  }
\end{table}

\begin{table}[t]
  \caption{Ablation studies on Multi-Stage Training Pipeline on KITTI val set.}
  \label{tab:ablation_multi_stage}
  \centering
  \resizebox{0.48\textwidth}{!}{
  \begin{tabular}{@{}ccc|ccc@{}}
  \toprule
  ~~Stage 1~~ & ~~Stage 2~~ & ~~Stage 3~~ & \begin{tabular}[c]{@{}c@{}}Car\\ (AP @ 3D-IoU 0.7)\end{tabular} & \begin{tabular}[c]{@{}c@{}}Pedestrian\\ (AP @ 3D-IoU 0.5)\end{tabular} & \begin{tabular}[c]{@{}c@{}}Cyclist\\ (AP @ 3D-IoU 0.5)\end{tabular} \\ \midrule
              &             & \checkmark  & 81.3                                                            & 60.4                                                                   & 77.9                                                                \\
              & \checkmark  & \checkmark  & 82.9                                                            & 61.7                                                                   & 79.1                                                                \\
  \checkmark  &             & \checkmark  & 83.5                                                            & 62.8                                                                   & 80.7                                                                \\
  \checkmark  & \checkmark  & \checkmark  & \textbf{85.6}                                                   & \textbf{64.1}                                                          & \textbf{81.2}                                                       \\ \bottomrule
  \end{tabular}
  }
\end{table}

We then perform ablation studies on the training strategy. The experiments are conducted under the sparsely-supervised setting on the KITTI dataset using Voxel-RCNN as the base detector. As shown in \cref{tab:ablation_prototype_training}, we evaluate the contributions of three key components: 
(1) the pseudo heatmap $H_{p}$ guides the model to focus on reliable pseudo-labeled regions, significantly improving performance over using only GT supervision; 
(2) the prototype similarity score map $H_{s}$ helps mine additional informative features, further boosting detection accuracy; 
(3) the contrastive loss encourages discriminative feature learning, leading to substantial gains when combined with $H_{s}$. The full combination of all three components yields the best results across all classes.

We also ablate the multi-stage training pipeline in \cref{tab:ablation_multi_stage}. Training with only Stage 3 (full prototype-based learning) results in suboptimal performance due to unstable prototype learning. Adding Stage 2 (prototype-based learning with GT only) improves stability and performance. Including Stage 1 (memory-based learning with GT only) further enhances prototype initialization and feature representation, leading to the best overall results. These findings validate the effectiveness of our multi-stage training approach in stabilizing and enhancing prototype-based learning.

For more ablation studies on hyper-parameters and training details, please refer to the supplementary material.

\section{Conclusion}

We presented SPL, a unified framework that effectively addresses unsupervised and sparsely-supervised 3D object detection. At its core, SPL introduces a semantic-aware pseudo-labeling mechanism that robustly generates 3D supervision from images and point clouds by integrating geometric cues and temporal consistency. Beyond label generation, the framework advances representation learning through a novel multi-stage prototype training strategy, which stabilizes feature mining and enhances discriminability without direct dependency on noisy pseudo-labels. Extensive evaluations on KITTI and nuScenes demonstrate that SPL significantly outperforms existing methods in both learning paradigms, validating its capability to learn effectively from minimal or no manual annotations. For future work, we will extend the pseudo-labeling to support more diverse object categories. We also plan to develop iterative optimization strategies, such as a self-training framework that progressively refines pseudo-labels and model parameters.

\section*{Acknowledgments}

This work is supported by the National Natural Science Foundation of China (Grant 62573287), and the Science and Technology Commission of Shanghai Municipality (Grant 20DZ2220400).

% \begin{thebibliography}{1}
\bibliographystyle{IEEEtran}
\bibliography{main}
% \end{thebibliography}

% \newpage

% \section{Biography Section}
% If you have an EPS/PDF photo (graphicx package needed), extra braces are
%  needed around the contents of the optional argument to biography to prevent
%  the LaTeX parser from getting confused when it sees the complicated
%  $\backslash${\tt{includegraphics}} command within an optional argument. (You can create
%  your own custom macro containing the $\backslash${\tt{includegraphics}} command to make things
%  simpler here.)
 
% \vspace{11pt}

% \bf{If you include a photo:}\vspace{-33pt}
% \begin{IEEEbiography}[{\includegraphics[width=1in,height=1.25in,clip,keepaspectratio]{fig1}}]{Michael Shell}
% Use $\backslash${\tt{begin\{IEEEbiography\}}} and then for the 1st argument use $\backslash${\tt{includegraphics}} to declare and link the author photo.
% Use the author name as the 3rd argument followed by the biography text.
% \end{IEEEbiography}

% \vspace{11pt}

% \bf{If you will not include a photo:}\vspace{-33pt}
% \begin{IEEEbiographynophoto}{John Doe}
% Use $\backslash${\tt{begin\{IEEEbiographynophoto\}}} and the author name as the argument followed by the biography text.
% \end{IEEEbiographynophoto}

\newpage

\appendix

\subsection{Implementation Details}

\begin{table*}[b]
  \caption{The detailed parameters for Pseudo-Label Generation.}
  \label{tab:notations}
  \centering
  \begingroup
  \renewcommand{\arraystretch}{1.5}
  \resizebox{1.0\textwidth}{!}{
  \begin{tabular}{@{}c|c|c@{}}
  \toprule
                                                                                             & Parameter Description                                                    & Value \\ \midrule
  Data Preprocessing                                                                         & Number of LiDAR frames aggregated in one process                         & KITTI: 1, nuScenes: 11 \\ \midrule
  \multirow{11}{*}{\begin{tabular}[c]{@{}c@{}} 3D Point \\ Label Generation \end{tabular}} & Kernel size when dilating the 2D object masks                            & $3 \times 3$ \\ \cline{2-3}
                                                                                             & Number of iterations when dilating the 2D object masks                   & \begin{tabular}[c]{@{}c@{}} $\sqrt{area_{mask} / 20}$, \\ $area_{mask}$ is the pixel area of the mask \end{tabular} \\ \cline{2-3}
                                                                                             & The prior height ranges $[h_{c, min}, h_{c, max}]$ for different classes & \begin{tabular}[c]{@{}c@{}} Pedestrian and Cyclist: 1.0-2.0 m, Car: 1.0-3.0 m, \\ Bus: 1.5-5.0 m, Truck: 1.5-5.0 m \end{tabular} \\ \cline{2-3}
                                                                                             & Parameter $\epsilon$ of DBSCAN                                           & Vehicle: 0.5 m, Pedestrian and Cyclist: 0.3 m \\ \cline{2-3}
                                                                                             & Parameter min\_samples of DBSCAN                                         & 4 \\ \cline{2-3}
                                                                                             & Search radius $r_{1}$                                                    & Vehicle: 0.25 m, Pedestrian and Cyclist: 0.15 m \\ \cline{2-3}
                                                                                             & Maximum radius $r_{2}$                                                   & Vehicle: 1.0 m, Pedestrian and Cyclist: 0.6 m \\ \cline{2-3}
                                                                                             & Number of neighboring points considered for KNN majority voting          & 10 \\ \cline{2-3}
                                                                                             & Minimum number of object points required to fit a 3D Bbox                & Vehicle: 20, Pedestrian: 10, Cyclist: 10 \\ \cline{2-3}
                                                                                             & Minimum distance to the box surface of SPR filtering                     & 0.2 m \\ \cline{2-3}
                                                                                             & Score threshold of SPR filtering                                         & 0.8 \\ \midrule
  \multirow{2}{*}{\begin{tabular}[c]{@{}c@{}} 3D Bbox \\ Label Generation \end{tabular}}  & Length, width and height ranges of 3D Bboxes for different classes       & \begin{tabular}[c]{@{}c@{}} Car: [2.0, 1.2, 1.2, 7.0, 2.0, 3.0], \\ Pedestrian and Cyclist: [0.25, 0.2, 0.5, 2.5, 1.0, 2.5], \\ Other Vehicle: [2.0, 1.2, 1.2, 20.0, 6.0, 6.0] \\ (Format: [Lmin, Wmin, Hmin, Lmax, Wmax, Hmax] m) \end{tabular} \\ \cline{2-3}
                                                                                             & Minimum speed for filtering Cyclist objects                              & 1 m/s \\ \bottomrule
  \end{tabular}
  }
  \endgroup
\end{table*}

In this section, we provide comprehensive implementation details of SPL, including pseudo-label generation and network training. Beyond listing parameters, we also clarify the design rationale behind key settings and how these settings are aligned with the proposed method in the main paper.

\subsubsection{Pseudo-Label Generation}

The detailed parameters of pseudo-label generation are summarized in \cref{tab:notations}. These settings are selected according to the characteristics of KITTI and nuScenes to balance pseudo-label precision and recall. Concretely, the mask-dilation and depth-range constraints are used to improve point association quality in crowded scenes, while DBSCAN/KNN-based refinement mitigates point misassignment and overlap ambiguity. The thresholds in SPR filtering and geometric size constraints are chosen to suppress implausible 3D boxes without over-pruning valid hard samples. In addition, the class-dependent search radii and minimum-point constraints are designed to reflect the density differences among Vehicle, Pedestrian, and Cyclist instances. Together, these choices provide a stable pseudo-label source for subsequent prototype-based training.

\subsubsection{Network and Training}

Following the baseline detector settings, we report only the additional parameters introduced by SPL and its multi-stage prototype learning strategy. The reported settings are directly tied to the core method components in \cref{sec:prototype_based_training_strategy,sec:multi_stage_training_pipeline}: prototype construction and update, pseudo-label-guided feature mining, contrastive objectives, and stage scheduling. For clarity, common detector defaults (e.g., standard voxelization and backbone channel settings) are intentionally omitted here because they are inherited from the baseline and are not specific to SPL.

\begin{table*}[t]
  \caption{The detailed parameters for SPL Network and Training.}
  \label{tab:spl_specific_training_params}
  \centering
  \begingroup
  \renewcommand{\arraystretch}{1.5}
  \resizebox{1.0\textwidth}{!}{
  \begin{tabular}{@{}c|c|c@{}}
  \toprule
  & Parameter Description & Value \\ \midrule
  \multirow{5}{*}{\begin{tabular}[c]{@{}c@{}}Prototype\\ Learning\end{tabular}}
  & Number of prototypes per class $K$ & 5 \\ \cline{2-3}
  & Prototype feature dimension $D$ & 64 \\ \cline{2-3}
  & Memory bank size in Stage 1 & 1000 \\ \cline{2-3}
  & Similarity threshold for mined features ($\tau_s$) & 0.9 \\ \cline{2-3}
  & Prototype momentum coefficient ($\alpha$) & 0.9 \\ \midrule

  \multirow{3}{*}{\begin{tabular}[c]{@{}c@{}}Prototype\\ Update\end{tabular}}
  & Maximum number of updates per prototype in one iteration
  & \begin{tabular}[c]{@{}c@{}}KITTI sparse: 50, KITTI unsup: 50 \\
    nuScenes sparse: 50, nuScenes unsup: 100\end{tabular} \\ \cline{2-3}
  & Prototype update source across stages
  & \begin{tabular}[c]{@{}c@{}}Stage 1: memory-bank pretraining and K-means initialization \\
    Stage 2: update with GT supervision only \\
    Stage 3: update with GT and mined pseudo-guided features\end{tabular} \\ \cline{2-3}
  & Number of prototypes sampled as background references & $C \times K$ \\ \midrule

  \multirow{4}{*}{\begin{tabular}[c]{@{}c@{}}Loss\\ Design\end{tabular}}
  & Additional losses introduced by SPL
  & \begin{tabular}[c]{@{}c@{}}Intra-class contrastive loss,\\ Inter-class contrastive loss\end{tabular} \\ \cline{2-3}
  & Heatmap classification and localization weights ($w_{cls}, w_{loc}$)
  & \begin{tabular}[c]{@{}c@{}}KITTI sparse: (1.0, 2.0), KITTI unsup: (1.0, 2.0) \\
    nuScenes sparse: (1.0, 0.25), nuScenes unsup: (1.0, 2.0)\end{tabular} \\ \cline{2-3}
  & Weight of intra-/inter-class contrastive losses ($\lambda_1,\lambda_2$)
  & \begin{tabular}[c]{@{}c@{}}KITTI sparse: (0.5, 1.0), KITTI unsup: (0.5, 1.0) \\
    nuScenes sparse: (0.25, 0.5), nuScenes unsup: (0.5, 1.0)\end{tabular} \\ \cline{2-3}
  & Contrastive temperature ($\tau_t$) & 1.0 \\ \midrule

  \multirow{3}{*}{\begin{tabular}[c]{@{}c@{}}Multi-stage\\ Training\end{tabular}}
  & Stage switching epochs (Stage 1 $\rightarrow$ Stage 2 $\rightarrow$ Stage 3)
  & \begin{tabular}[c]{@{}c@{}}KITTI sparse: 0/10/20, KITTI unsup: 0/10/20 \\
    nuScenes sparse: 0/5/10, nuScenes unsup: 0/5/10\end{tabular} \\ \cline{2-3}
  & Late-stage disabling of GT sampling augmentation
  & \begin{tabular}[c]{@{}c@{}}KITTI sparse and unsup: GT sampling kept throughout training \\
    nuScenes sparse: last 10 epochs, nuScenes unsup: last 2 epochs\end{tabular} \\ \cline{2-3}
  & Total epochs under multi-stage training
  & \begin{tabular}[c]{@{}c@{}}KITTI sparse: 80, KITTI unsup: 80 \\
    nuScenes sparse: 30, nuScenes unsup: 22\end{tabular} \\ \bottomrule
  \end{tabular}
  }
  \endgroup
\end{table*}

As summarized in \cref{tab:spl_specific_training_params}, the parameter groups are organized according to the training logic of SPL. The \textit{Prototype Learning} block controls representation capacity and matching strictness (e.g., $K$, $D$, and $\tau_s$). The \textit{Prototype Update} block governs update stability and prevents noisy assignments from dominating the prototype bank, especially in early training. The \textit{Loss Design} block defines the balance between the base detection objective ($w_{cls}, w_{loc}$) and the intra-/inter-class contrastive terms, which is crucial for jointly optimizing localization quality and feature discrimination. Finally, the \textit{Multi-stage Training} block specifies when to transition from memory-based initialization to full pseudo-guided prototype learning and when to disable strong augmentation in late epochs for more stable convergence.

Across the four settings, most SPL-specific hyper-parameters remain consistent, indicating good transferability of the framework. The main differences appear in (1) detection and contrastive loss weights, (2) maximum prototype update counts, and (3) stage/augmentation schedules. These differences reflect dataset-level characteristics: nuScenes sparse supervision adopts lighter localization and contrastive weights to account for class heterogeneity and broader scene variability, while nuScenes unsupervised increases the prototype update cap to improve adaptability under weaker direct supervision.

\subsection{Qualitative Results}

\begin{figure*}[b]
  \centering
  \includegraphics[width=18cm]{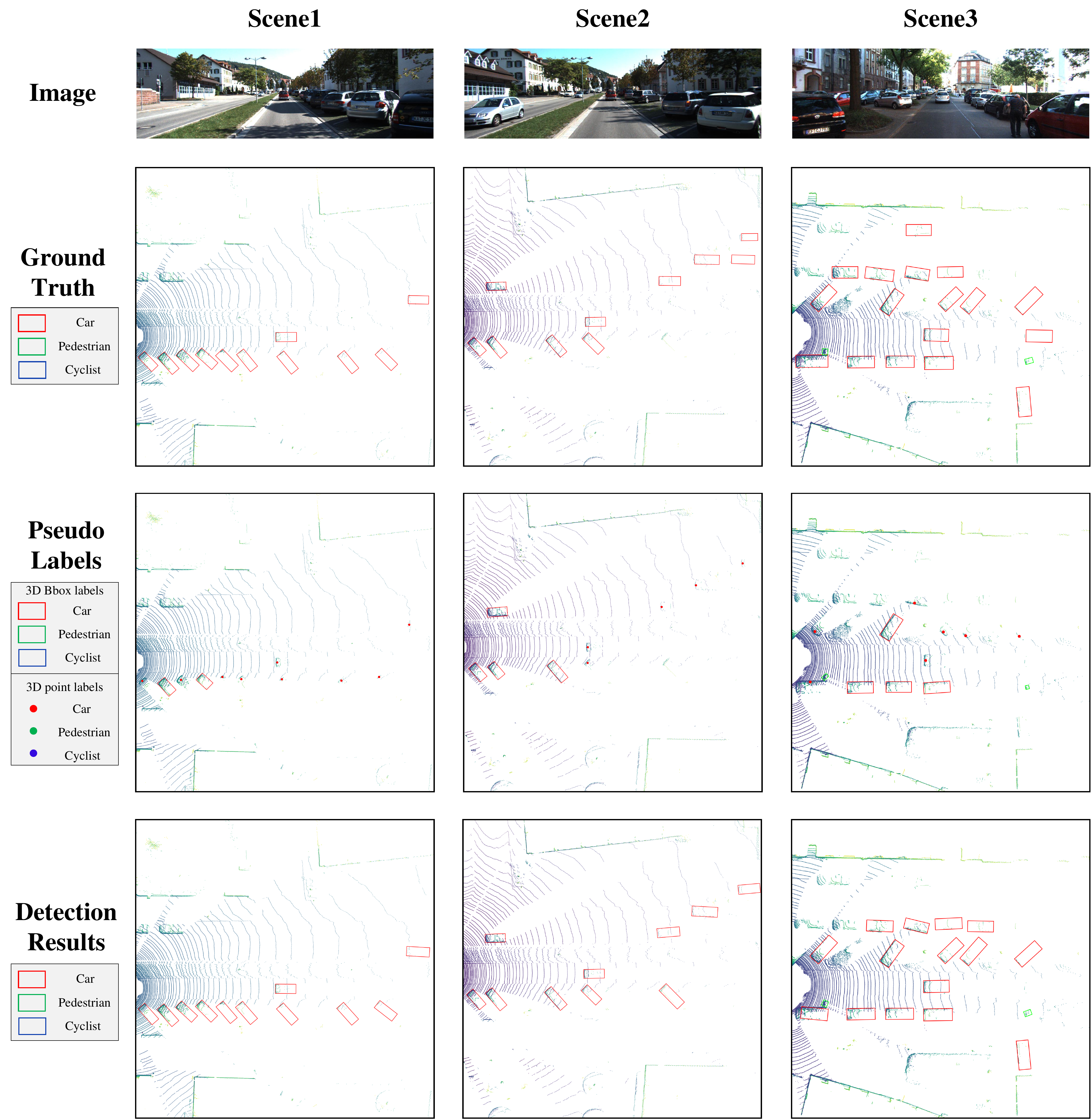}
  \caption{The visualization of ground truth annotations, pseudo labels, and detection results of our SPL method on KITTI dataset. (Part 1)}
  \label{fig:detection_results1}
\end{figure*}

\begin{figure*}[b]
  \centering
  \includegraphics[width=18cm]{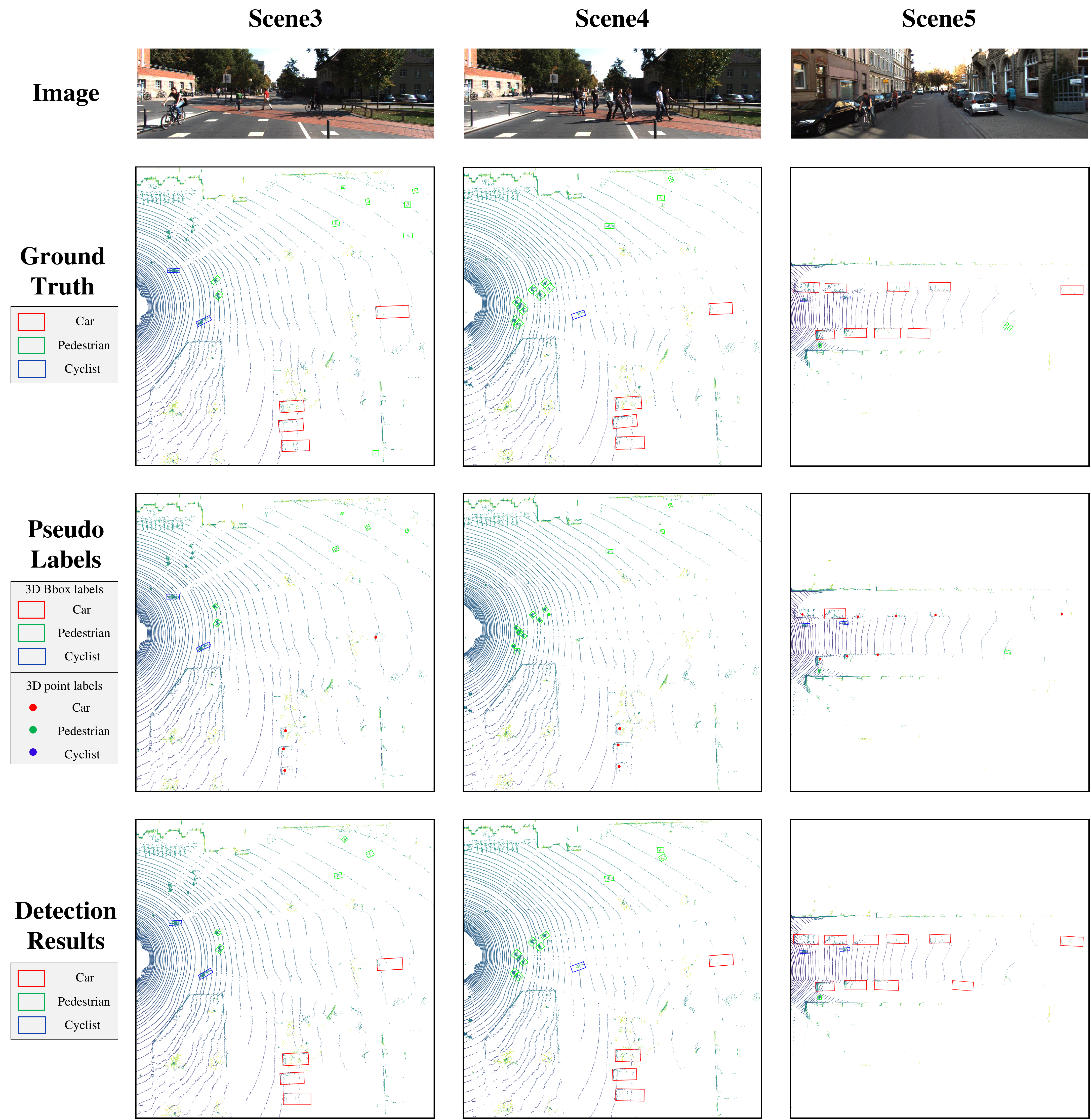}
  \caption{The visualization of ground truth annotations, pseudo labels, and detection results of our SPL method on KITTI dataset. (Part 2)}
  \label{fig:detection_results2}
\end{figure*}

In this section, we present qualitative results of SPL on KITTI. For each scene, we visualize ground-truth annotations, generated pseudo labels, and final detection results under sparse supervision. As shown in \cref{fig:detection_results1} and \cref{fig:detection_results2}, SPL produces reliable pseudo labels for most objects, and many Pedestrian/Cyclist instances obtain accurate 3D bounding box pseudo labels. For distant Car instances with very sparse points, fitting precise 3D boxes remains difficult; therefore, these instances are mainly represented by 3D point pseudo labels.

A notable observation is that model predictions after training are substantially cleaner and more complete than the initial pseudo labels. This behavior is consistent with our design: pseudo labels are used as priors for feature mining rather than hard supervisory targets, and prototype-guided contrastive learning helps recover discriminative structure from noisy or incomplete annotations. Qualitatively, this leads to fewer background confusions, better object extent estimation for medium-density instances, and improved category consistency across nearby frames. These observations support the quantitative gains reported in the main paper and further validate the effectiveness of SPL in both representation learning and detection refinement.

\vfill

\end{document}